\documentclass{article}

\PassOptionsToPackage{numbers, compress}{natbib}
\usepackage[final]{neurips_data_2024}

\usepackage[utf8]{inputenc} 
\usepackage[T1]{fontenc}    
\usepackage{hyperref}       
\usepackage{url}            
\usepackage{booktabs}       
\usepackage{amsfonts}       
\usepackage{nicefrac}       
\usepackage{microtype}      
\usepackage{xcolor}         
\usepackage{xspace}
\usepackage{graphicx}
\usepackage{enumitem}
\usepackage{pifont}
\usepackage{subcaption}
\usepackage{tcolorbox}

\hypersetup{
  colorlinks,
  citecolor=blue!90,
  linkcolor=blue!70,
  urlcolor=blue!70
}

\newcommand{\arena}{\textsc{GenAI-Arena}\xspace}

\title{GenAI Arena: An Open Evaluation Platform for Generative Models}

\author{
Dongfu Jiang$^*$ \quad 
Max Ku$^*$ \quad 
Tianle Li$^*$ \and
\textbf{
\quad Yuansheng Ni
\quad Shizhuo Sun
\quad Rongqi Fan
\quad Wenhu Chen} \vspace{0.4em} \\
University of Waterloo \\
\small{\texttt{ 
\{dongfu.jiang, m3ku, t29li, wenhuchen\}@uwaterloo.ca
}} \vspace{0.4em}
\\
\small{\texttt{\url{https://hf.co/spaces/TIGER-Lab/GenAI-Arena}}}
\vspace{-2em}
}

\begin{document}

\maketitle

\setlist[itemize]{leftmargin=0.15in}

\begin{abstract}
Generative AI has made remarkable strides to revolutionize fields such as image and video generation. These advancements are driven by innovative algorithms, architecture, and data. However, the rapid proliferation of generative models has highlighted a critical gap: the absence of trustworthy evaluation metrics. Current automatic assessments such as FID, CLIP, FVD, etc often fail to capture the nuanced quality and user satisfaction associated with generative outputs. This paper proposes an open platform \arena to evaluate different image and video generative models, where users can actively participate in evaluating these models. By leveraging collective user feedback and votes,  \arena aims to provide a more democratic and accurate measure of model performance. It covers three tasks of text-to-image generation, text-to-video generation, and image editing respectively. Currently, we cover a total of 35 open-source generative models. \arena has been operating for seven months, amassing over 9000 votes from the community. We describe our platform, analyze the data, and explain the statistical methods for ranking the models. To further promote the research in building model-based evaluation metrics, we release a cleaned version of our preference data for the three tasks, namely GenAI-Bench. We prompt the existing multi-modal models like Gemini, and GPT-4o to mimic human voting. We compute the accuracy by comparing the model voting with the human voting to understand their judging abilities. Our results show existing multimodal models are still lagging in assessing the generated visual content, even the best model GPT-4o only achieves an average accuracy of $49.19\%$ across the three generative tasks. Open-source MLLMs perform even worse due to the lack of instruction-following and reasoning ability in complex vision scenarios.
\end{abstract}

\begingroup
\setlength{\intextsep}{5pt}
\begin{figure}[!ht]
    \centering
    \vspace{-1em}
    \includegraphics[width=0.83\linewidth]{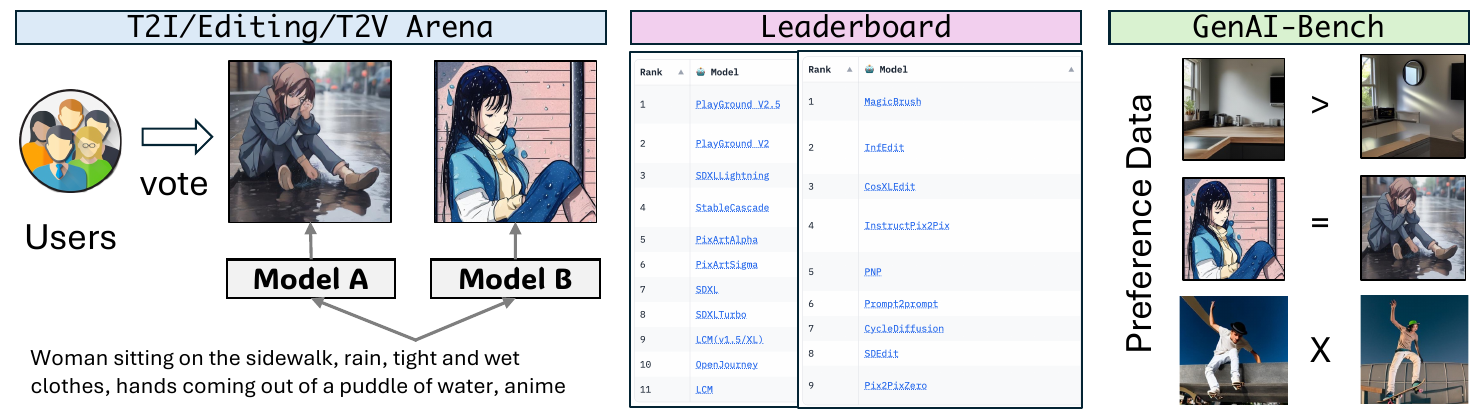}
    \caption{GenAI Arena contains three components: (1) text-to-image, text-to-video and image editing arena, which accept community voting to obtain the preference pairs. (2) The leaderboard utilizes the preference pairs to calculate elo ranking for all the evaluated models. (3) We further release GenAI-Bench to judge different multimodal LLM judges. }
    \label{fig:teaser}
\end{figure}
\endgroup

\section{Introduction}
\label{intro}
Image generation and manipulation technologies have seen rapid advancements, leading to their widespread application across various domains such as creating stunning artwork~\cite{nichol2022glide, saharia2022photorealistic, zhang2023adding, ho2022imagen}, enhancing visual content~\cite{brooks2022instructpix2pix, li2023dreamedit}, and aiding in medical imaging~\cite{zhang2024lefusion, chen2024generalizable}. Despite these advancements, navigating through the multitude of available models and assessing their performance remains a challenging task \cite{Ramesh2022HierarchicalTI}. Traditional evaluation metrics like PSNR, SSIM~\cite{ssim_1284395}, LPIPS~\cite{zhang2018perceptual}, and FID~\cite{heuselfid2017}, while valuable, offer very specific insights into precise aspects of visual content generation. However, these metrics often fall short in providing a comprehensive assessment of overall model performance, especially when considering subjective qualities like aesthetics and user satisfaction~\cite{otani2023verifiable}. 

To address these challenges, we introduce GenAI-Arena—a novel platform designed to enable fair evaluation. Inspired by successful implementations in other domains~\cite{zheng2023judging, tts-arena}, GenAI-Arena offers a dynamic and interactive platform where users can generate images, compare them side-by-side, and vote for their preferred models. Such a platform not only simplifies the process of comparing different models but also provides a ranking system that reflects human preferences, thereby offering a more holistic evaluation of model capabilities. To our knowledge, GenAI-Arena is the first evaluation platform with comprehensive evaluation capabilities across multiple properties. Unlike other platforms, it supports a wide range of tasks across text-to-image generation, text-guided image editing, and text-to-video generation, along with a public voting process to ensure labeling transparency. The votes are utilized to access the evaluation ability of Multimodal Large Language Model (MLLM) evaluators. Table~\ref{tab:baselines} shows our platform excels in its versatility and transparency.

Since February 11th, 2024, we have collected over 9000 votes for three multimodal generative tasks. We constructed leaderboards for each task with these votes, identifying the state-of-the-art models as PlayGround V2.5, MagicBrush, and StableVideoDiffusion, respectively (until Oct 24th, 2024). Detailed analyses based on the votes are presented. For example, our plotted winning fraction heatmaps reveal that while the Elo rating system is generally effective, it can be biased by imbalances between "easy games" and "hard games". We also performed several case studies for qualitative analysis, demonstrating that users can provide preference votes from multiple evaluation aspects, which help distinguish subtle differences between the outputs and upload high-quality votes for Elo rating computation.

Automatically assessing the quality of generated visual content is a challenging problem for several reasons: (1) images and videos have many different aspects like visual quality, consistency, alignment, artifacts, etc. Such a multi-faceted nature makes the evaluation intrinsically difficult. (2) the supervised data is relatively scarce on the web. In our work, we release the user voting data as GenAI-Bench to enable further development in this field. Specifically, we calculate the accuracy between different image/video auto-raters (i.e. MLLM judges like GPT-4o, Gemini, etc.) with user preference to understand their judging abilities. Our results show that even the best MLLM, GPT-4o achieves at most $49.19\%$ accuracy compared with human preference.

\definecolor{darkgreen}{rgb}{0.0, 0.8, 0.0}
\definecolor{darkred}{rgb}{0.8, 0.0, 0.0}
\begin{table}[!h]
\centering
\vspace{-1em}
\caption{Comparison with different evaluation platforms on different properties.}
\small
\def\arraystretch{1.0}
\setlength\tabcolsep{2 pt}
\scalebox{0.85}{
\begin{tabular}{lcccccc}
\toprule
\textbf{Platform}  & \textbf{\begin{tabular}[c]{@{}c@{}}Text-To-Image\\ Generation\end{tabular}} & \textbf{\begin{tabular}[c]{@{}c@{}}Text-Guided\\ Image Editing\end{tabular}} & \textbf{\begin{tabular}[c]{@{}c@{}}Text-To-Video\\ Generation\end{tabular}} & \textbf{\begin{tabular}[c]{@{}c@{}}Human Label\\ Transparency\end{tabular}} & \textbf{\begin{tabular}[c]{@{}c@{}} Open/Public \\ Voting Process\end{tabular}} & \textbf{\begin{tabular}[c]{@{}c@{}} Judging \\ MLLM judge\end{tabular}} \\
\midrule
T2I-CompBench~\cite{huang2023t2icompbench} & \textcolor{darkgreen}{\ding{51}} & \textcolor{darkred}{\ding{55}} & \textcolor{darkred}{\ding{55}} & \textcolor{darkred}{\ding{55}} & \textcolor{darkred}{\ding{55}} & \textcolor{darkred}{\ding{55}} \\ 
HEIM~\cite{lee2024holistic} & \textcolor{darkgreen}{\ding{51}} & \textcolor{darkred}{\ding{55}} & \textcolor{darkred}{\ding{55}} & \textcolor{darkgreen}{\ding{51}} & \textcolor{darkred}{\ding{55}} & \textcolor{darkred}{\ding{55}}  \\ 
ImagenHub~\cite{ku2024imagenhub} & \textcolor{darkgreen}{\ding{51}} & \textcolor{darkgreen}{\ding{51}} & \textcolor{darkred}{\ding{55}} & \textcolor{darkgreen}{\ding{51}} & \textcolor{darkred}{\ding{55}} & \textcolor{darkred}{\ding{55}}\\ 
VBench~\cite{huang2023vbench} & \textcolor{darkred}{\ding{55}} & \textcolor{darkred}{\ding{55}} & \textcolor{darkgreen}{\ding{51}} & \textcolor{darkgreen}{\ding{51}} & \textcolor{darkred}{\ding{55}} & \textcolor{darkred}{\ding{55}}\\ 
EvalCrafter~\cite{liu2023evalcrafter} & \textcolor{darkred}{\ding{55}} & \textcolor{darkred}{\ding{55}} & \textcolor{darkgreen}{\ding{51}} & \textcolor{darkgreen}{\ding{51}} & \textcolor{darkred}{\ding{55}} & \textcolor{darkred}{\ding{55}} \\ 
\midrule
\textbf{\arena} & \textcolor{darkgreen}{\ding{51}} & \textcolor{darkgreen}{\ding{51}} & \textcolor{darkgreen}{\ding{51}} & \textcolor{darkgreen}{\ding{51}} & \textcolor{darkgreen}{\ding{51}} & \textcolor{darkgreen}{\ding{51}}\\ 
\bottomrule
\end{tabular}}
\label{tab:baselines}
\end{table}

To summarize, our work's contributions include:
\begin{itemize}[topsep=0pt,itemsep=0.05ex]
    \item GenAI-Arena, the first open platform to rank multi-modal generative AI based on user preferences.
    \item Discussion and case studies of collected user votes, showing the reliability of GenAI-Arena.
    \item GenAI-Bench, a public benchmark for judging MLLM's evaluation ability for generative tasks. 
\end{itemize}

\section{Related Work}
\label{rela_work}

\subsection{Generative AI Evaluation Metrics}
Numerous methods have been proposed to evaluate the performance of multi-modal generative models in various aspects. In the context of image generation, CLIPScore~\cite{hessel2021clipscore} is proposed to measure the text-alignment of an image and a text through computing the cosine similarity of the two embeddings from CLIP~\cite{radford2021learning_CLIP}. IS~\cite{NIPS2016_8a3363ab} and FID~\cite{heuselfid2017} measure image fidelity by computing a distance function between real and synthesized data distributions. PSNR, SSIM~\cite{ssim_1284395} assess the image similarity. LPIPS~\cite{zhang2018perceptual} and the follow-up works~\cite{fu2024dreamsim, ghazanfari2023lipsim} measure the perceptual similarity of images. More recent works leverage the Multimodal Large Language Model (MLLM) as a judge. T2I-CompBench~\cite{huang2023t2icompbench} proposed the use of miniGPT4~\cite{zhu2023minigpt} to evaluate compositional text-to-image generation task. TIFA~\cite{hu2023tifa} further adapted visual question answering to compute scores for the text-to-image generation task. VIEScore~\cite{ku2023viescore} leveraged MLLMs as a unified metric across image generation and editing tasks, reporting that MLLM has great potential in replacing human judges.

Metrics in similar fashions are also proposed for the video domain. For example, FVD~\cite{unterthiner2018towards} measures the coherence shifts and quality in frames. CLIPSIM~\cite{radford2021learning_CLIP} utilizes an image-text similarity model to assess the similarity between video frames and text. VBench~\cite{huang2023vbench} and EvalCrafter~\cite{liu2023evalcrafter} also proposed different metrics for evaluating different aspects of the video generation task. However, these automatic metrics still lag compared with human preferences, achieving low correlation and thus giving doubts to their reliability.

\subsection{Generative AI Evaluation Platforms}
While auto-metric focuses on evaluating a single model's performance, evaluation platforms aim to systematically rank a group of models. Recently, several benchmark suites have been developed to comprehensively assess generative AI models. For image generation, T2ICompBench~\cite{huang2023t2icompbench} evaluates compositional text-to-image generation tasks, while HEIM~\cite{lee2024holistic} offers a holistic evaluation framework that measures text-to-image tasks across multiple dimensions, including safety and toxicity. Similarly, ImagenHub~\cite{ku2024imagenhub} evaluates text-to-image, image editing, and other prevalent image generation tasks in a unified benchmark suite. For video generation, VBench~\cite{huang2023vbench} and EvalCrafter~\cite{liu2023evalcrafter} provide structured evaluation approaches ensuring rigorous assessment. Despite their functionality, these benchmarks rely on model-based evaluation metrics, which are less reliable than human evaluation.

To address this issue, variable model arenas have been developed to collect direct human preferences for ranking models. Chatbot Arena by LMsys~\cite{chiang2024chatbot} is the pioneering platform in this regard, setting the standard for evaluation. Subsequent efforts have led to the creation of arenas for vision-language models~\cite{xu2023lvlm}, TTS models~\cite{tts-arena}, and tokenizers~\cite{tokenizer_arena}. However, there is no existing arena for generative AI models. To fill this gap, we propose GenAI-Arena as a complementary solution in this field.

\section{GenAI-Arena: Design and Implementation}
\label{GenAI-Arena}

\begingroup
\setlength{\intextsep}{0pt}
\begin{figure}[!ht]
    \centering
    \includegraphics[width=0.95\linewidth]{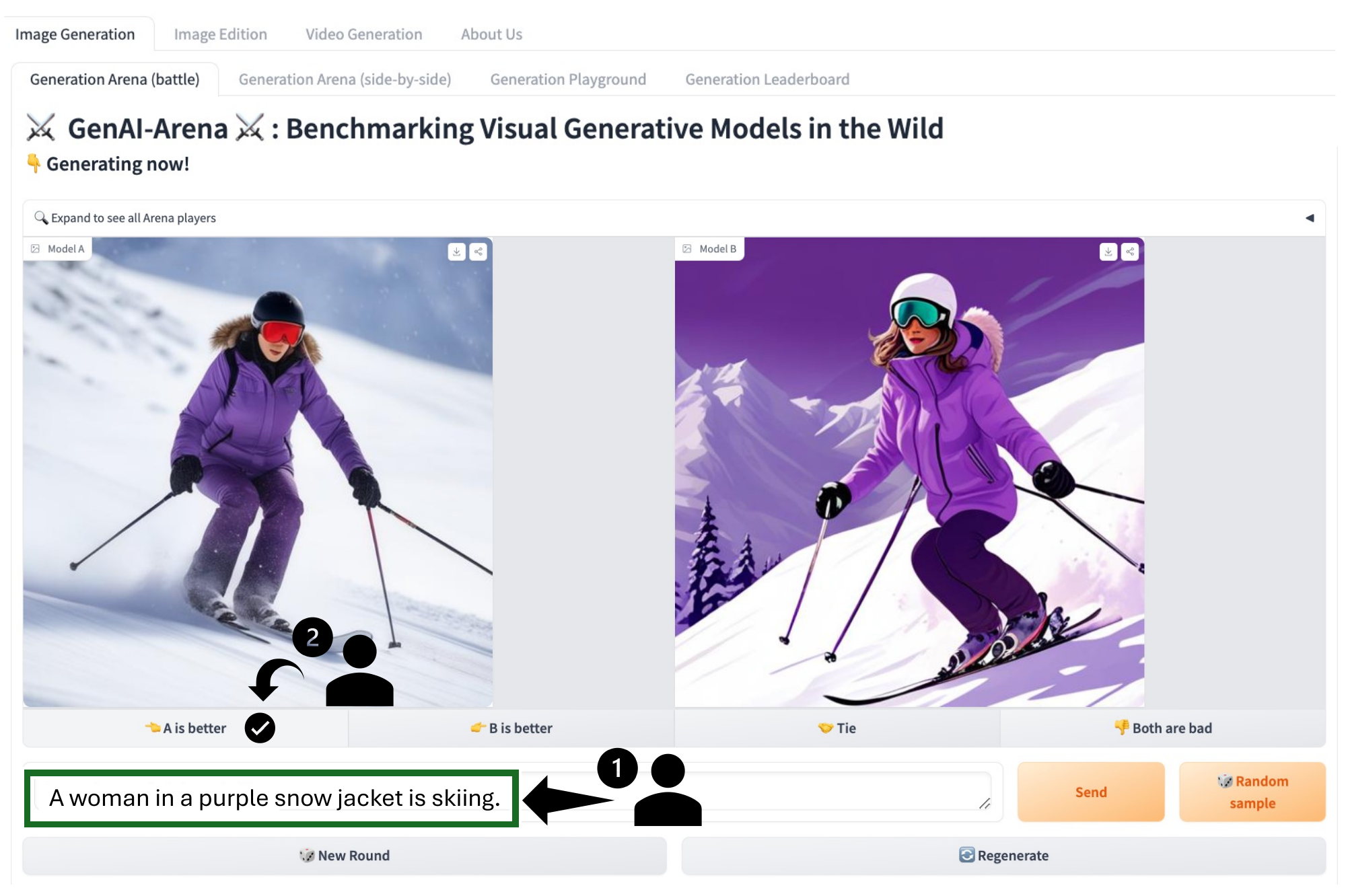}
    \caption{GenAI Arena User Voting Interface. }
    \label{fig:demo}
\end{figure}
\endgroup


\subsection{Design}
GenAI-Arena is designed to offer an intuitive and comprehensive evaluation platform for generative models, facilitating user interaction and participation. The platform is structured around three primary tasks: text-to-image generation, image edition, and text-to-video generation. Each task is supported by a set of features that include an anonymous and a non-anonyumous battle playground, a direct generation tab, and a leaderboard as shown in Figure~\ref{fig:demo} . These features are designed to cater to both casual users and researchers, ensuring a democratic and accurate assessment of model performance.

\paragraph{Standardized Inference}
To ensure a fair comparison between different models, we ported the highly dispersed codebase from the existing works and then standardized them into a unified format. During inference, we fixed the hyper-parameters and the prompt format to prevent per-instance prompt or hyper-parameter tuning, which makes the inference of different models fair and reproducible. Following ImagenHub~\cite{ku2024imagenhub}, we build the new library of VideoGenHub (details in  ~\autoref{sec:videogenhub}), which aims to standardize the inference procedure for different text-to-video and image-to-video models. We find the best hyper-parameters of these models to ensure their highest performance.  

\paragraph{Voting Rules}
The anonymous battle section is designed to ensure unbiased voting and accurate evaluation of generative models. The rules for this section are as follows:
\begin{enumerate}[leftmargin=*, itemsep=0ex]
\item Users input a prompt, which is then used to generate outputs from two anonymous models within the same category of task. 
\item The generated outputs from the two anonymous models are presented side-by-side for comparison.
\item Users can vote based on their preference using the options: 1) left is better; 2) right is better; 3) tie; 4) both are bad. These four options are being used to calculate Elo ranking.
\item Once the user has made their decision, they click the Vote button to submit their vote. It is important to ensure that the identity of the models remains anonymous throughout the process. Votes will not be counted if the model identity is revealed during the interaction.
\end{enumerate}

\subsection{Model Integration}
In GenAI-Arena, we incorporate a diverse array of state-of-the-art generative models, covering a broad range of generative tasks including text-to-image generation, image edition, and text-to-video generation. To ensure comprehensive evaluations, the platform includes models that employ diverse underlying technologies, such as different types of architectures, training paradigms, training data and acceleration techniques. These variations can offer insights to understand these factors rigorously.

\paragraph{Text-to-Image Generation}
In~\autoref{tab:t2i}, we list all the included text-to-image generation models. For example, SDXL, SDXL-Turbo, and SDXL-Lightning are all derived based on SDXL~\cite{Podell2023SDXLIL}, while SDXL-Turbo~\cite{Sauer2023AdversarialDD} and SDXL-Lightning~\cite{Lin2024SDXLLightningPA} adopt different distillation method. We also include diffusion transformer models~\cite{peebles2023scalable} like PixArt-$\alpha$ and PixArt-$\sigma$. Playground V2 and Playground V2.5 are based on SDXL architecture, but trained by Playground.ai from scratch with an internal dataset. We have also included the latest released HunyuanDiT~\cite{li2024hunyuandit}, FLUX.1-dev~\cite{blackforestlabs_flux}, FLUX.1-schnell~\cite{blackforestlabs_flux}.

\begin{table}[!ht]
\small
\centering
\vspace{-1em}
\caption{The overview of all text-to-image generation models.}
\scalebox{0.9}{
\begin{tabular}{lcccc}
\toprule
Model                                   & Size              & Method                                         & Resolution   & \#Steps \\
\midrule
OpenJourney~\cite{openjourney}          & 1B                & SD-2.1 + MidJourney Dataset                    & 512x512      & 50      \\
LCM~\cite{Luo2023LatentCM}              & 1B                & SD-2.1 + Consistency Distillation              & 512x512      & 4       \\
SDXL~\cite{Podell2023SDXLIL}                & 3.5B              & Latent Diffusion Model (LDM)                        & 1K$\times$1K & 50      \\
SDXL-Turbo~\cite{Sauer2023AdversarialDD}    & 3.5B        & LDM + Distillation          & 1K$\times$1K & 1       \\
SDXL-Lightning~\cite{Lin2024SDXLLightningPA}& 3.5B        & LDM + Distillation          & 1K$\times$1K & 4       \\
PixArt-$\alpha$~\cite{Chen2023PixArtFT}     & 0.6B        & Diffusion Transformer (DiT)                             & 1K$\times$1K & 50      \\
PixArt-$\sigma$~\cite{Chen2024PixArtWT}     & 0.6B        & DiT + Weak-to-Strong            & 4K$\times$4K & 50      \\
StableCascade~\cite{pernias2023wurstchen}   & 1.5B + 3.6B & Würstchen                          & 1K$\times$1K & 20+10   \\
Playground V2~\cite{playground-v2}          & 3.5B        & LDM                                  & 1K$\times$1K & 50      \\
Playground V2.5~\cite{Li2024PlaygroundVT}   & 3.5B        & LDM                            & 1K$\times$1K & 50      \\
FLUX.1-dev~\cite{blackforestlabs_flux} & 12B & Guidance-distilled DiT + Flow Matching & 1K$\times$1K & 20\\
FLUX.1-schnell~\cite{blackforestlabs_flux} & 12B & Timestep-distilled DiT + Flow Matching & 1K$\times$1K & 4\\
Kolors~\cite{kwai_kolors} & 2.6B & LDM + ChatGLM3 & 1K$\times$1K & 50 \\
HunyuanDiT~\cite{li2024hunyuandit} & 1.5B & DiT + multilingual text encoder & 1K$\times$1K & 50\\ 
Stable Diffusion 3~\cite{stabilityai_stablediffusion3} & 8B & Multimodal DiT & 1K$\times$1K & 50\\
AuraFlow~\cite{huggingface_auraflow} & 6.8B & Flow-based Model & 1K$\times$1K & 50\\
\bottomrule
\end{tabular}}
\label{tab:t2i}
\end{table}

\begin{table}[!ht]
\small
\centering
\vspace{-1em}
\caption{Overview of all the image editing models. }
\scalebox{0.95}{
\begin{tabular}{lccc}
\toprule
Model                                 & Trained? & Method                                                          & Runtime \\
\midrule
Pix2PixZero~\cite{parmar2023zero}     & Zero-shot & Editing Direction Discovery + Attention Control                & 21s     \\
SDEdit~\cite{meng2021sdedit}          & Zero-shot & Iteratively Denoising through SDE                              & 13s     \\
CycleDiffusion~\cite{cyclediffusion}  & Zero-shot &  Reconstructable Encoder for Stochastic DPMs                   & 9s      \\
Prompt2Prompt~\cite{Hertz2022PrompttoPromptIE}    & Zero-shot   &  Prompt-based Cross-attention Control            & 120s    \\
PnP~\cite{tumanyan2023plug}           & Zero-shot &  Feature and Self-attention Injection                          & 120s    \\
InfEdit~\cite{xu2023infedit}          & Zero-shot &  Consistent Model + Uni-Attention Control                      & 5s      \\
InstructPix2Pix~\cite{brooks2022instructpix2pix}  & Trained   &  Instruction-based Fine-tuning with Synthetic Data & 12s     \\
MagicBrush~\cite{zhang2023magicbrush} & Trained   &  Instruction-based Fine-tuning with Annotated Data             & 12s     \\
CosXLEdit~\cite{CosXL}                & Trained   &  Cosine-Continuous EDM VPred schedule                          & 50s     \\
\bottomrule
\end{tabular}}
\label{tab:i2i}
\end{table}
\begin{table}[!ht]
\small
\centering
\vspace{-1em}
\caption{Overview of all text-to-video generation models.}
\resizebox{0.895\linewidth}{!}{
\begin{tabular}{lcccccc}
\toprule
Model                                               & Base            & Len           & FPS & Dataset         & Resolution   & \#Steps \\
\midrule
AnimateDiff~\cite{guo2023animatediff}               & SD-1.5          & 2s            & 8   & WebVid10M                & 512$\times$512   & 25      \\
AnimateDiff-Turbo~\cite{guo2023animatediff}         & SD-1.5          & 2s            & 8   & WebVid10M                &  512$\times$512  & 4       \\
ModelScope~\cite{Wang2023ModelScopeTT}              & SD-1.5          & 2s            & 8   & WebVid10M        &  256$\times$256  &  50       \\
LaVie~\cite{wang2023lavie}                          & SD-1.5          & 2s            & 8   & Vimeo25M                 &  320$\times$512  & 50      \\
StableVideoDiffusion~\cite{Blattmann2023StableVD}   & SD-2.1          & 2.5s          & 10  & LVD-500M                 & 576$\times$1024   & 20      \\
VideoCrafter2~\cite{Chen2024VideoCrafter2OD}        & SD-2.1          & 2s            & 16  & WebVid10M        & 320$\times$512   & 50      \\
T2V-Turbo~\cite{Li2024T2VTurboBT}                   & VideoCrafter2   & 2s            & 8   & WebVid10M                &  320$\times$512  & 4       \\
OpenSora~\cite{opensora}                            & Pixart-$\alpha$ & 2s            & 16  & WebVid10M        &  320$\times$512  & 50      \\
OpenSora v1.2~\cite{opensora}                            & Pixart-$\alpha$ & 2s            & 16  & WebVid10M        &  320$\times$512  & 50      \\
CogVideoX-2B~\cite{yang2024cogvideox} & DiT & 2s & 8 & 35M videos + 2B images & 480$\times$720 & 50 \\
\bottomrule
\end{tabular}
}
\label{tab:t2v}
\end{table}

\paragraph{Text-guided Image Editing}
In~\autoref{tab:i2i}, we list all the image editing models and approaches. Some of them are plug-and-play approaches without requiring any training, like Pix2PixZero~\cite{parmar2023zero}, InfEdit~\cite{xu2023infedit}, SDEdit~\cite{meng2021sdedit}, etc. These methods can be applied to a broad range of diffusion models. Some of the models like PnP~\cite{tumanyan2023plug} and Prompt2Prompt~\cite{Hertz2022PrompttoPromptIE} require DDIM inversion, which takes much longer time than the other approaches. We also include specialized trained image editing models like InstructP2P~\cite{brooks2022instructpix2pix}, MagicBrush~\cite{zhang2023magicbrush} and CosXLEdit~\cite{CosXL}.

\paragraph{Text-to-Video Generation}
In~\autoref{tab:t2v}, we list all the text-to-video generation models. We include different types of models. For example, AnimateDiff~\cite{guo2023animatediff}, ModelScope~\cite{Wang2023ModelScopeTT}, Lavie~\cite{wang2023lavie} are initialized from SD-1.5 and continue trained by injecting a motion layer to capture the temporal relation between frames. In contrast, StableVideoDiffusion~\cite{Blattmann2023StableVD} and VideoCrafter2~\cite{chen2024videocrafter2} are iniialized from SD-2.1. Besides these models, we also include OpenSora~\cite{opensora}, which utilizes a Sora-like diffusion transformer~\cite{peebles2023scalable} architecture for joint space-time attention.

\subsection{Elo Rating System}

\paragraph{Online Elo Rating}
The Elo rating system models the probability of player $i$ winning against player $j$, based on their current ratings, $R_i$ and $R_j$ respectively, where $i, j \in N$. We define a binary outcome $Y_{ij}$ for each comparison between player $i$ and player $j$, where $Y_{ij} = 1$ if player $i$ wins and $Y_{ij} = 0$ otherwise. The logistic probability is formulated as:
\begin{equation}
P(Y_{ij}=1) = \frac{1}{1 + 10^{ (R_j - R_i) / \alpha}}
\end{equation}
where $\alpha = 400$ for Elo rating computation. After each match, a player's rating is updated using the formula: 
\begin{equation}
    R^{\prime}_i = R_i + K \times (S(i,j) - E(i,j))
\end{equation}
where $S(i,j)$ is the actual match outcome, $S(i,j)=1$ for a win $S(i,j)=0.5$ for a tie, and $S(i,j)=0$ for a loss, and $E(i,j) = P(Y_{ij} = 1)$. K is 

For example, given a model's Elo rating as 1200 and the other model's elo rating as 1100, then the estimated probability of the first model winning will be $\frac{1}{1+10^{(1100-1200)/400}}\approx0.64$. In this way, we can have a direct understanding of the elo rating's meaning. This mapping from absolute number to the pairwise winning rate of two models gives a more straightforward understanding of the meaning of elo rating score. 

Another design logic behind the Elo rating is that a higher-rated player should gain fewer points if they win a lower-rated player, but lose more if they lose the game, whereas the lower-rated player experiences the opposite. In this way, the order of a specific set of matches will significantly affect the final computed Elo rating, as the player's Elo rating and the rating gain of each match are both changing dynamically. This online Elo rating system might be good for real-world competitions, where players usually have less than 100 competitions a year. However the arena for AI models usually comes with thousands of votes (competitions), and the quality of votes is not ensured. Thus, it's necessary to acquire an order-consistent and more stable elo rating. To do this, we follow Chatbot Arena~\cite{Chiang2024ChatbotAA} to adopt the Bradley–Terry model~\cite{bradley1952rank} for a statistically estimated elo rating.

\paragraph{Bradley–Terry Model Estimation}
The Bradley–Terry (BT) model~\cite{bradley1952rank} estimates Elo ratings using logistic regression and maximum likelihood estimation (MLE). Suppose there are $N$ players and we have a series of pairwise comparisons, where $W_{ij}$ is the number of times player $i$ has won against player $j$. The log-likelihood function for all pairwise comparisons is written as:
\begin{equation}
\mathcal{L}(\mathbf{R}) = \sum_{i,j \in N, i \neq j} \left( W_{ij} \log P(Y_{ij} = 1) \right)
= \sum_{i,j \in N, i \neq j} \left( W_{ij} \log \frac{1}{1 + 10^{ (R_j - R_i) / \alpha}} \right)
\end{equation}
where $\mathbf{R} = \{R_1, \ldots, R_N\}$ represents the Elo ratings of each player. The Bradley–Terry model provides a stable statistical estimation of the players' ratings by consistently incorporating all pairwise comparisons, thus overcoming the limitations of direct Elo computation in online settings.

Since the BT model does not account for ties, we first duplicate all the votes, then allocate half of the "tie" votes to the scenario where model $i$ wins ($Y_{ij} = 1$) and the other half to the scenario where model $j$ wins ($Y_{ij} = 0$) in practice. We model the solver to be a logistic regression model and solve it via the \texttt{LogisticRegression} model from \texttt{sklearn} for the solving.

\paragraph{Confidence Interval}
To further investigate the variance of the estimated Elo rating, we use the "sandwich" standard errors described in ~\citet{huber1967behavior}. That is, for each round, we record the estimated Elo rating based on the same number of battles sampled from the previous round. This process continues for 100 rounds. We select the lowest sampled elo rating as the lower bound of the confidence interval, and the highest sampled elo rating as the upper bound of the elo rating. 

\paragraph{Selection of battle pair}
With a limited number of games, choosing which two players to match up is a crucial issue. The simplest approach, which we currently use, is to randomly select two players. However, this can introduce bias, with some models getting significantly more matches than others. A vote-aware selection system that increases the probability of selecting less-played models and lowers it for more-played ones is needed, and we plan to explore this in future Arena improvements.

\subsection{GenAI-Museum}
\label{sec:genai-museum}
Current GenAI-Arena runs the model on the Hugging Face Zero GPU system~\cite{ZeroGPU}. As shown in ~\autoref{tab:i2i}, the time for a single generative inference usually ranges from 5 to 120 seconds. Unlike the auto-regression language model, where inference acceleration techniques like VLLM~\cite{kwon2023efficient}, SGLang~\cite{zheng2023efficiently} generate responses in less than a second, diffusion model community does not have such powerful infrastructure. Therefore, pre-computation becomes a necessary way to mitigate computational overhead and streamline user interaction.

To achieve this, we serve GenAI-Museum as a pre-computed data pool comprising various inputs from existing datasets or user collection, along with each model's output. Based on this, a \texttt{"Random Sample"} button shown in ~\autoref{fig:demo} is additionally implemented to facilitate the random generation of prompts and the immediate retrieval of corresponding images or videos. This functionality operates by sending requests to our deployed GenAI-Museum every time \texttt{"Random Sample"} button is hit, receiving input and two random model's pre-computed outputs. In this way, we save the computation time on the GPU, enable users to do instant comparisons and votes on the UI, and balance the votes for each unique input so we gradually collect votes for a full combination of all models. The input prompts were sampled from ImagenHub~\cite{ku2024imagenhub} and VBench~\cite{huang2023vbench}. To prevent the bias in the prompt distribution, we also periodically update the input prompts with the lastest collected real-world human votes. We make sure every prompt is filtered via NSFW detector before adding them.

\section{Benchmarks and Results Discussion}
\label{bench}
\subsection{Arena Leaderboard}

We report our leaderboard at the time of paper publishing in~\autoref{tab:leaderboard}. For image generation, we collected 6300 votes in total. The currently top-1 model is Playground V2.5, released by Playground.ai, which follows the same architecture as SDXL but is trained with a private dataset.
In contrast, SDXL only ranks in the thirteenth position, lagging significantly behind. Such finding highlights the importance of the training dataset. 
StableCascade is ranked in the sixth place in the leaderboard, which utilizes a highly efficient cascade architecture to lower the training cost. According to Würstchen~\cite{pernias2023wurstchen}, StableCascade only requires a 10\% training cost of SD-2.1, yet it can beat SDXL significantly on our leaderboard. This highlights the importance of the diffusion architecture to achieve strong performance. For image editing, a total of 1154 votes have been collected. MagicBrush, InFEdit, CosXLEdit, and InstructPix2Pix ranked higher as they can perform localized editing on images. PNP preserves the structure with feature injections, thus limiting the edit variety. The older methods such as Prompt-to-Prompt, CycleDiffusion, SDEdit, and Pix2PixZero, frequently result in completely different images during editing despite the high-quality images, which explains the lower ranking of these models. For text-to-video, there is a total of 2024 votes. StableVideoDiffusion leads with the highest Elo score, suggesting it is the most effective model. Close behind, CogVideoX-2B ranks second. The following VideoCrafter2 and AnimateDiff have very close elo scores, showing nearly equivalent capabilities. LaVie, OpenSora, ModelScope, and AnimateDiff-Turbo follow with decreasing scores, indicating progressively lower performance.

\begin{table}[!t]
    \small
    \centering
    \vspace{-2em}
    \caption{GenAI-Arena Leaderboards. \\(Last updated on Oct 24th, 2024)}
    \scalebox{0.99}{
    \begin{minipage}[b]{0.32\textwidth}
        \centering
        \subcaption{Text-to-Image (Top-10)}
        \scalebox{0.85}{
        \begin{tabular}{lcc}
            \toprule
            Model & Elo & 95\% CI \\
            \midrule
            PlayGround V2.5 & 1122 & +19/-20 \\
            FLUX.1-dev    & 1114 & +45/-42 \\
            FLUX.1-schnell & 1085 & +43/-46 \\
            Playground V2 & 1072 & +18/-22 \\
            Kolors & 1069 & +32/-39 \\
            StableCascade & 1057 & +17/-21 \\
            HunyuanDiT & 1030 & +25/-27 \\
            PixArt-$\alpha$ & 1020 & +17/-19 \\
            SDXL-Lightning & 1020 & +17/-15 \\
            PixArt-$\sigma$ & 1019 & +22/-20 \\
            \bottomrule
        \end{tabular}}
        \label{tab:t2i_ranking}
    \end{minipage}
    \hfill
    \begin{minipage}[b]{0.30\textwidth}
        \centering
        \subcaption{Image Editing}
        \scalebox{0.85}{
        \begin{tabular}{lcc}
            \toprule
            Model & Elo & 95\% CI \\
            \midrule
            MagicBrush & 1108 & +32/-28 \\
            InfEdit & 1075 & +26/-32 \\
            CosXLEdit & 1066 & +31/-29 \\
            InstructPix2Pix & 1038 & +32/-24 \\
            PNP & 998 & +35/-34 \\
            Prompt2prompt & 988 & +26/-23 \\
            CycleDiffusion & 943 & +26/-26 \\
            SDEdit & 924 & +24/-23 \\
            Pix2PixZero & 858 & +25/-30 \\
            \\
            \bottomrule
        \end{tabular}}
        \label{tab:i2i_ranking}
    \end{minipage}
    \hfill
    \begin{minipage}[b]{0.33\textwidth}
        \centering
        \subcaption{Text-to-Video}
        \scalebox{0.85}{
        \begin{tabular}{lcc}
        \toprule
        Model & Elo & 95\% CI \\
        \midrule
        StableVideoDiffusion & 1148 & +31/-28 \\
        CogVideoX-2B & 1106 & +71/-71 \\
        T2V-Turbo & 1085 & +36/-32 \\
        VideoCrafter2 & 1068 & +20/-22\\
        AnimateDiff & 1068 & +25/-21 \\
        LaVie & 996 & +25/-21 \\
        OpenSora & 912 & +23/-23 \\
        OpenSora v1.2 & 894 & +54/-71 \\
        ModelScope & 862 & +25/-22 \\
        AnimateDiff-Turbo & 861 & +22/-20 \\
        \bottomrule
        \end{tabular}}
        \label{tab:t2v_ranking}
    \end{minipage}}
    \hfill
    \label{tab:leaderboard}
\end{table}

\begin{figure}[!h]
    \centering
    \vspace{-1em}
    \begin{subfigure}[b]{0.32\textwidth}
        \centering
        \caption{Text-to-Image}
        \includegraphics[width=\textwidth]{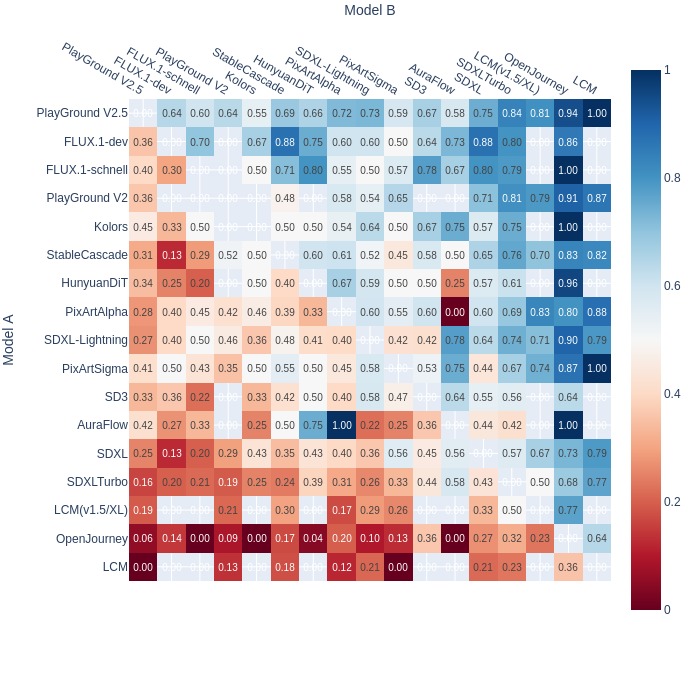}
        \label{fig:t2i_winfrac_heatmap}
    \end{subfigure}
    \hfill
    \begin{subfigure}[b]{0.32\textwidth}
        \centering
        \caption{Image Editing}
        \includegraphics[width=\textwidth]{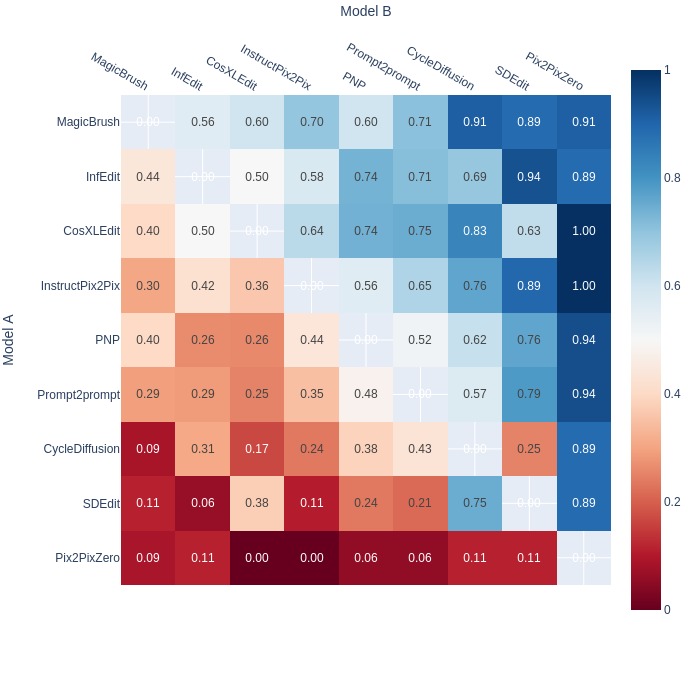}
        \label{fig:i2i_winfrac_heatmap}
    \end{subfigure}
    \hfill
    \begin{subfigure}[b]{0.32\textwidth}
        \centering
        \caption{Text-to-Video}
        \includegraphics[width=\textwidth]{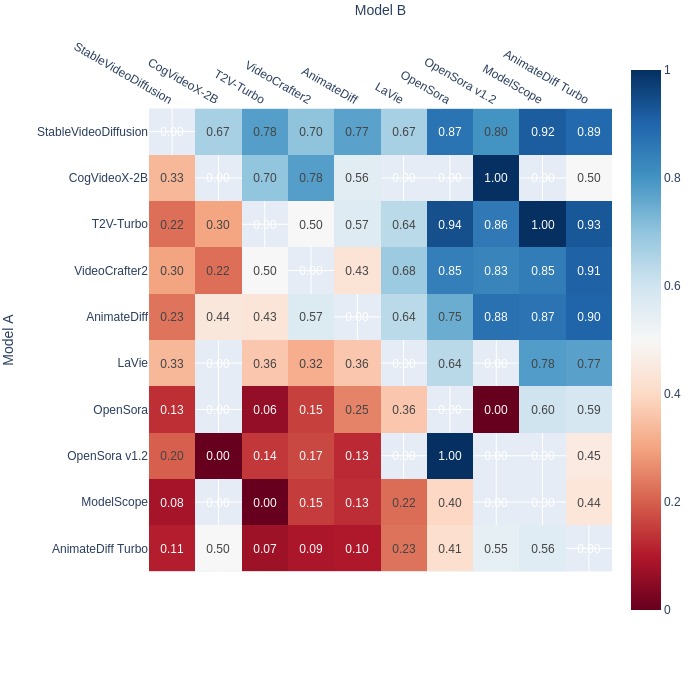}
        \label{fig:t2v_winfrac_heatmap}
    \end{subfigure}
    \vspace{-2em}
    \caption{Winning fraction heatmap of different models for the three tasks in GenAI-Arena}
    \vspace{1em}
    \label{fig:winfrac_heatmap}
    \begin{subfigure}[b]{0.32\textwidth}
        \centering
        \caption{Text-to-Image}
        \includegraphics[width=\textwidth]{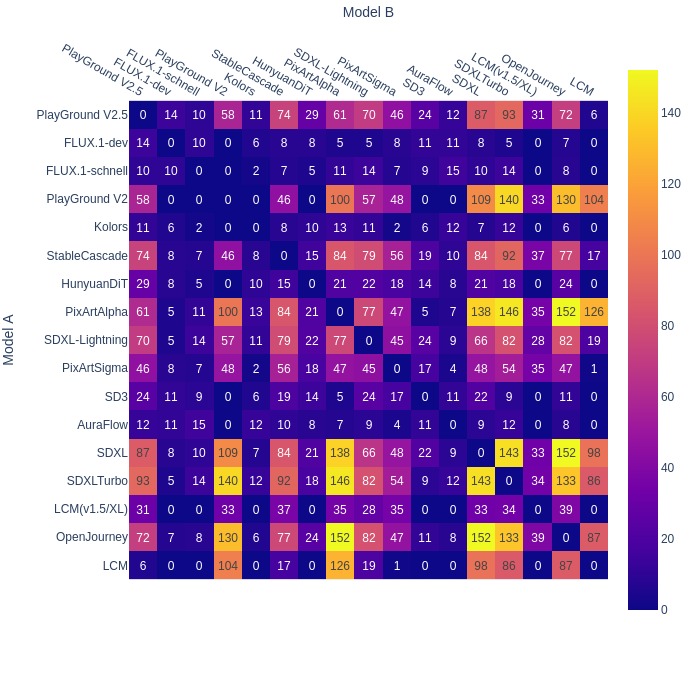}
        \label{fig:t2i_count_heatmap}
    \end{subfigure}
    \hfill
    \begin{subfigure}[b]{0.32\textwidth}
        \centering
        \caption{Image Editing}
        \includegraphics[width=\textwidth]{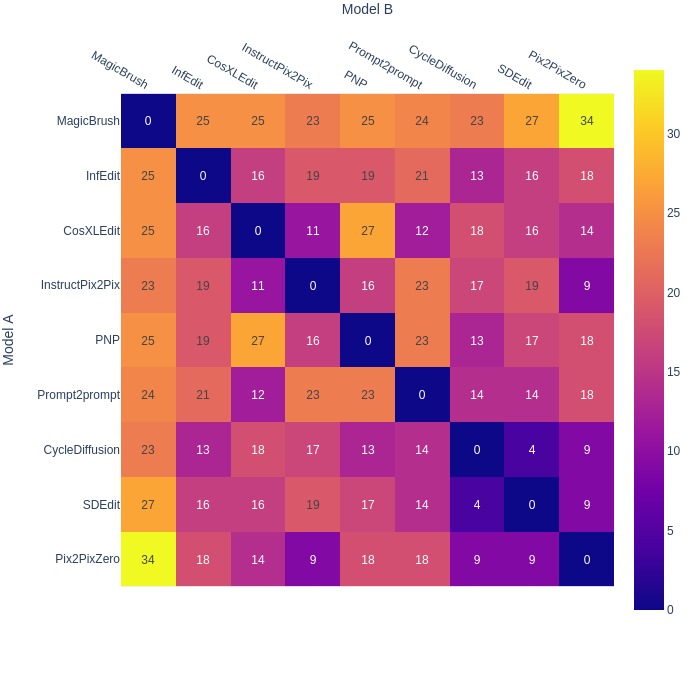}
        \label{fig:i2i_count_heatmap}
    \end{subfigure}
    \hfill
    \begin{subfigure}[b]{0.32\textwidth}
        \centering
        \caption{Text-to-Video}
        \includegraphics[width=\textwidth]{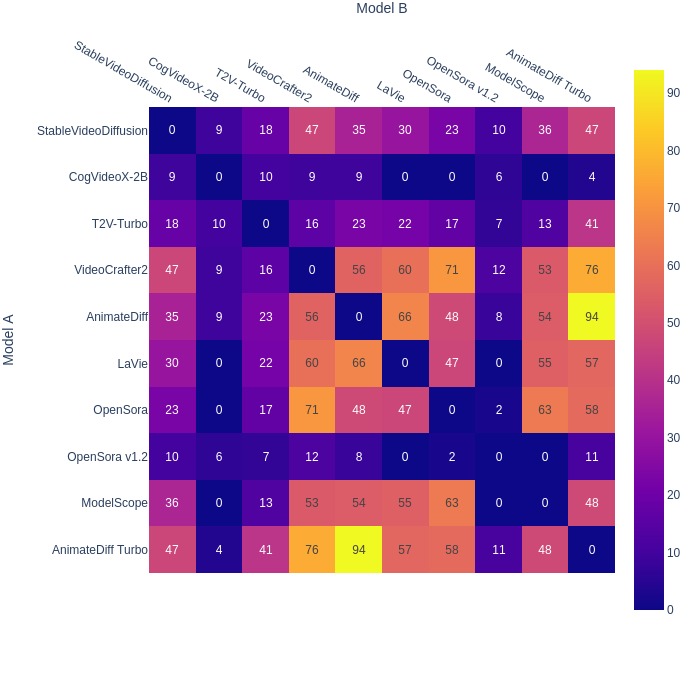}
        \label{fig:t2v_count_heatmap}
    \end{subfigure}
    \vspace{-2em}
    \caption{Battle count heatmap of different models for the three tasks in GenAI-Arena (without Ties)}
    \vspace{1em}
    \label{fig:count_heatmap}
    \begin{subfigure}[b]{0.32\textwidth}
        \centering
        \caption{Text-to-Image}
        \includegraphics[width=\textwidth]{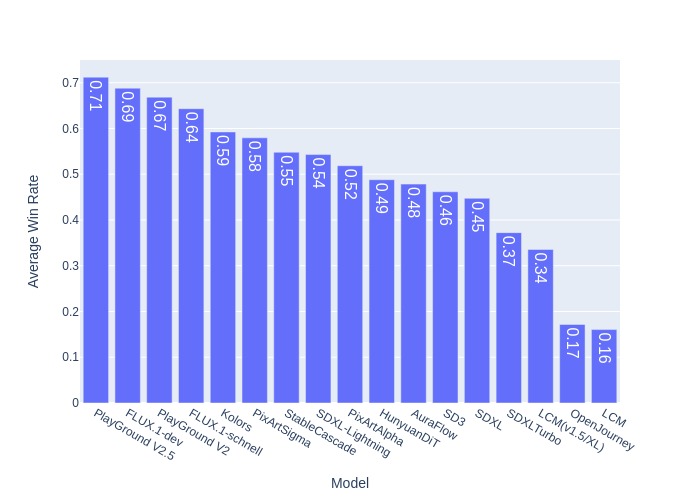}
        \label{fig:t2i_avg_winrate}
    \end{subfigure}
    \hfill
    \begin{subfigure}[b]{0.32\textwidth}
        \centering
        \caption{Image Editing}
        \includegraphics[width=\textwidth]{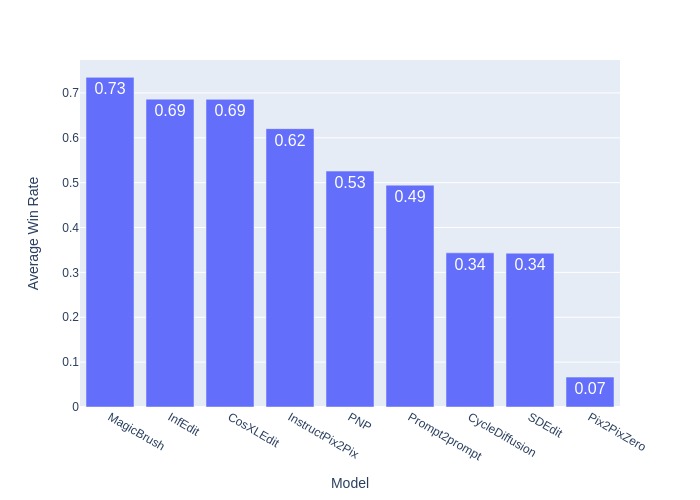}
        \label{fig:i2i_avg_winrate}
    \end{subfigure}
    \hfill
    \begin{subfigure}[b]{0.32\textwidth}
        \centering
        \caption{Text-to-Video}
        \includegraphics[width=\textwidth]{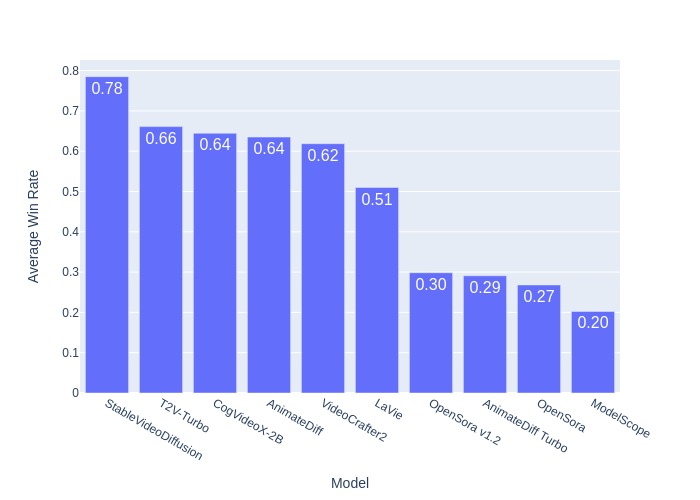}
        \label{fig:t2v_avg_winrate}
    \end{subfigure}
    \vspace{-1em}
    \caption{Average Win Rate Against All Other Models (Assuming Uniform Sampling and No Ties)}
    \label{fig:avg_winrate}
    \vspace{-1em}
\end{figure}
\subsection{Discussion and Insights}
\paragraph{Winning Fraction and Elo Rating}
We visualize the winning fraction heatmap in ~\autoref{fig:winfrac_heatmap}, where each cell represents the actual winning fraction of Model A over Model B. The models are ordered by their Elo rating in the heatmap. Horizontally across each row, the winning fraction of Model A increases as the Elo rating of Model B decreases, demonstrating the effectiveness of the Elo rating system in ranking different models.

Specific cells in the heatmap reveal notable findings. For instance, although PlayGround 2.5 achieves the state-of-the-art (SOTA) Elo rating in the Text-to-Image task, its winning fraction over PixArt-$\sigma$ is only $0.58$, which is below 60\%. The higher Elo rating of T2V-Turbo might be due to our Arena collecting more votes from "easy games" with low-ranked models and fewer from "harder games" with high-ranked models. For example, the number of battles between PlayGround V2.5 and SDXL-Turbo ($93$) is way more than PlayGround V2.5 with other models (around $50$) in ~\autoref{fig:count_heatmap}.

These anomalies highlight potential drawbacks of the Elo rating system: (1) a reliable and robust Elo rating requires a large amount of voting data, and (2) the estimated Elo rating may be biased by the imbalance between "easy games" and "harder games," as they carry similar weight in the estimation.

As shown in ~\autoref{fig:avg_winrate}, we observe that the average win rates of the top-ranked models are all quite similar, none exceeding 80\%. This indicates that there is no dominant, highly powerful model in text-to-image, image editing, or text-to-video generation at this time. The community is still awaiting a "ChatGPT moment"—the release of a breakthrough model with transformative capabilities.

\paragraph{Quality assessment of collected human votes}
Since our arena users come from different backgrounds and have different preferences, we conduct an expert review on a small set of sampled human vote to ensure there are no severe quality issues of our collected votes. We let different authors review 50 items for each set. A total of 350 items from our GenAI-Bench are evaluated. During the annotations, we skipped those bad items due to NSFW or technical issues, and we finally collected 303 valid evaluations. For each vote, 3 available labels for provided for annotating:
\begin{itemize}
    \item Clearly Reasonable Vote: This vote will be clearly agreed by most of the people. 
    \item Vague Vote: The current vote makes sense. But it's also reasonable if other vote is selected.
    \item Wrong Vote: This vote will be clearly disagreed by most of the people. 
\end{itemize}
\begin{table}[!h]
    \small
    \centering
    \caption{Expert Review for 350 sampled human votes}
    \scalebox{0.99}{
    \begin{minipage}[c]{0.4\textwidth}
        \centering
        \subcaption{Distribution of Valid Votes}
        \scalebox{0.80}{
        \begin{tabular}{lccc}
            \toprule
            \# Valid votes & \# NSFW & \# Tech issue & Total \\
            \midrule
            \textbf{303} & 17 & 30 & 350 \\
            \textbf{86.57}\% & 4.86\% & 8.57\% & 100\% \\
            \bottomrule
        \end{tabular}}
        \label{tab:dist_valid_votes}
    \end{minipage}
    \hfill
    \begin{minipage}[c]{0.45\textwidth}
        \centering
        \subcaption{Distribution of quality labels}
        \scalebox{0.85}{
        \begin{tabular}{cccc}
            \toprule
            \# Clearly Reasonable Vote & \# Vague Vote & \# Wrong Vote & Total \\
            \midrule
            \textbf{231} & 51 & 21 & 303 \\
            \textbf{76.24\%} & 16.83\% & 6.93\% & 100\% \\
            \bottomrule
        \end{tabular}}
        \label{tab:dist_quality_labels}
    \end{minipage}
    }
    \hfill
    \label{tab:vote_quality_review}
\end{table}

We report the distribution of valid votes in ~\autoref{tab:dist_valid_votes}, and find that 86.57\% of the votes are valid without NSFW issues. Among these valid votes, about 76.24\% of the votes are clearly reasonable votes and 93.07\% of the votes are either clearly reasonable or vaguely reasonable, as shown in ~\autoref{tab:dist_quality_labels}. We believe this shows the reliability of our preference data.

\begin{figure}
    \centering
    \includegraphics[width=\textwidth]{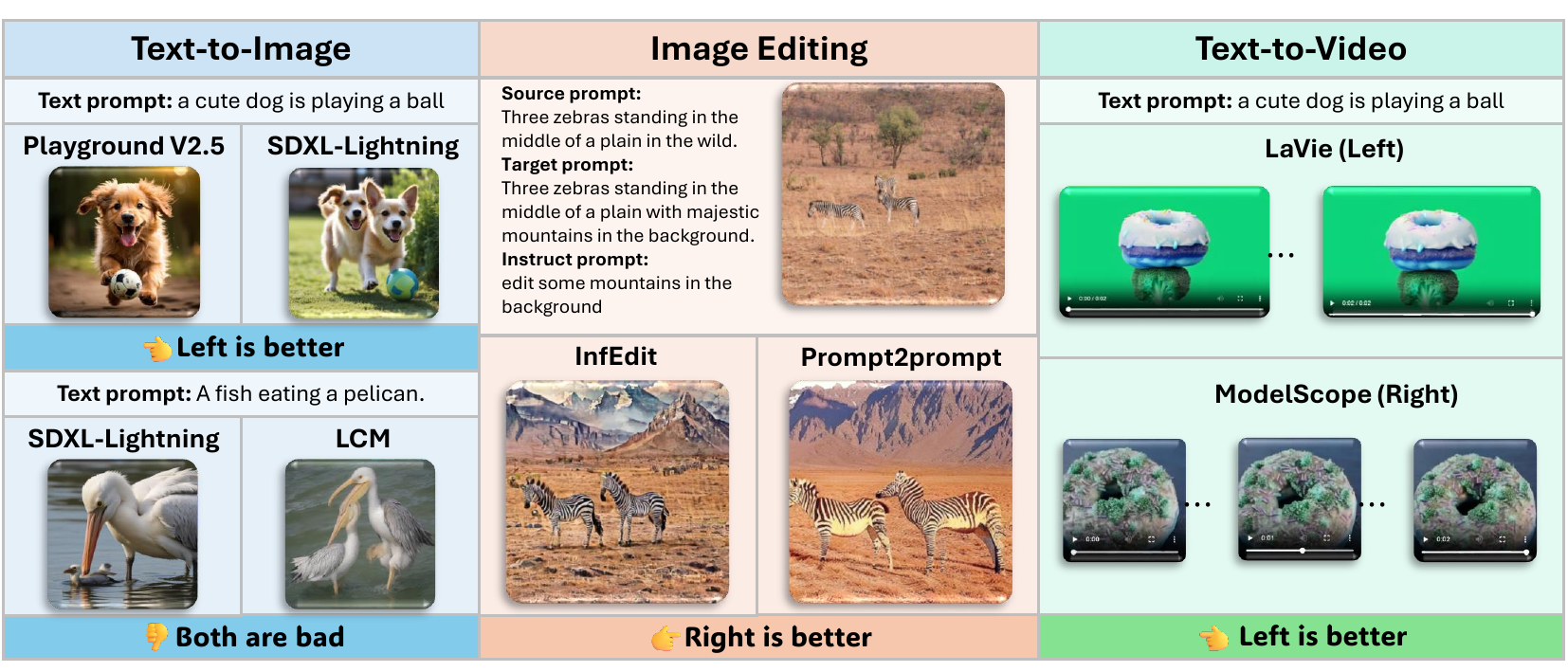}
    \caption{Example of votes from users on the GenAI-Arena for the three generative tasks}
    \label{fig:case_study}
\end{figure}
\paragraph{Case Study}
We present case studies in ~\autoref{fig:case_study}, showcasing the votes collected for three generative tasks. These cases demonstrate that GenAI-Arena users can provide high-quality votes, even for the most advanced models. For instance, in the text-to-image task, the image generated by PlayGround V2.5 was preferred over that of SDXL-Lightning for the prompt "a cute dog is playing with a ball," as the latter depicted two dogs instead of one. Users can clearly distinguish and vote based on the quality of the outputs, even when both models complete the task. In the image editing task, the edited image from Prompt2Prompt appeared more natural than the one from InfEdit, leading users to make a definitive vote. Similarly, votes collected for the text-to-video task were also of high quality.

\section{GenAI-Bench}
\label{sec:genai_bench}
\subsection{Dataset}
We applied Llama Guard~\cite{Inan2023LlamaGL} as an NSFW filter to ensure that the user input prompt is appropriate for a wide range of audiences and protects users of the benchmark from exposure to potentially harmful or offensive content. In the text-to-image generation task, we collect 4.3k anonymous votes in total and there are 1.7k votes left after filtering for the safe content. We observe a large amount of the prompt is filtered out due to sexual content, which takes up 85.6\% of the abandoned data. In the text-guided image editing task, we collect 1.1k votes from users before filtering. After applying Llama Guard, there are 0.9k votes for the image edition being released. In this task,  87.5\% of the unsafe inputs contain violent crimes, and the other 12.5\% is filtered out resulting from sex-related crimes. For text-to-video generation task, our platform collects 1.2k votes before post-processing. After cleaning it with the NSFW filter, we release the remaining 1.1k votes. All of the unsafe data abandoned in this task is due to the sexual content. We released the current version of GenAI-Bench\footnote{\url{https://huggingface.co/datasets/TIGER-Lab/GenAI-Bench}} on the HuggingFace Dataset website, with an MIT license to allow the reuse with or without modification.


\subsection{GenAI-Bench Leaderboard}
\label{subsec:pairwise_acc}
To construct the GenAI-Bench leaderboard, we propmt MLLMs to output preference labels of AI generated contents, where templates are defined in ~\autoref{subsec:genai_bench_prompt_templates}. Specifically, We selected MLLMs including GPT-4o~\cite{openai2023gpt4}, Gemini-1.5-Pro~\cite{reid2024gemini}, Idefics2~\cite{laurençon2024matters}, etc., and ask them to output 4 labels: ``\textsc{[[A>B]]}'', ``\textsc{[[B>A]]}'', ``\textsc{[[A=B=Good]]}'', and ``\textsc{[[A=B=Bad]]}''. We then compare them with actual human preference labels collected through the GenAI-Arena using the exact match metric. As shown in Table~\ref{tab:genaibench_results}, open-source model still lag behind close-source MLLMs such as GPT-4o and Gemini, indicating a lack of generalization ability in vision reasoning of open-source MLLMs. We also tried models including Fuyu~\cite{fuyu-8b}, Kosmos-2~\cite{Peng2023Kosmos2GM}, Otter~\cite{Li2023OtterAM}, Mantis~\cite{jiang2024mantis}, etc., but found that they cannot follow the instruction well to output reasonable labels.

\begin{table}[h]
\centering
\caption{GenAI-Bench leaderboard designed to benchmark MLLMs's ability in judging the quality of AI generative contents by comparing with human preferences. Numbers are accuracy (\%).}
\small
\def\arraystretch{1.0}
\setlength\tabcolsep{2 pt}
\begin{tabular}{lcccc}
\toprule
Model                    & Image Generation & Image Editing & Video Generation & Average \\
\midrule
Random                   & 25.36            & 25.90         & 25.16            & 25.47   \\
\midrule
Idefics1~\cite{laurencon2023obelics}                 & 0.81             & 5.66          & 0.19             & 2.22    \\
InstructBLIP~\cite{brooks2022instructpix2pix}            & 3.11             & 19.80         & 3.74             & 8.89    \\
QwenVL~\cite{Bai2023QwenVLAF}                   & 26.63            & 14.91         & 2.15             & 14.56   \\
CogVLM~\cite{Wang2023CogVLMVE}                   & 29.34            & 0.00          & 24.60            & 17.98   \\
VideoLLaVA~\cite{Lin2023VideoLLaVALU}               & 37.75            & 26.66         & 0.00             & 21.47   \\
BLIP-2~\cite{li2023blip}       & 26.34            & 26.01         & 16.93            & 23.09   \\
MiniCPM-V-2.5~\cite{Hu2024MiniCPMUT}    & 37.81            & 25.24         & 6.55             & 23.20   \\
LLaVA-1.6-7B~\cite{liu2024llavanext}             & 22.65            & 25.35         & 21.70            & 23.24   \\
Idefics2~\cite{laurençon2024matters}                 & 42.25            & 27.31         & 16.46            & 28.67   \\
LLaVA-1.5-7B~\cite{Liu2023ImprovedBW}             & 37.00            & 26.12         & 30.40            & 31.17   \\
Gemini-1.5-Pro~\cite{reid2024gemini}           & 44.67            & \textbf{55.93}& 46.21            & 48.94   \\
GPT-4o~\cite{openai2023gpt4}                   & \textbf{45.59}   & 53.54         & \textbf{48.46}   & \textbf{49.19}   \\
\bottomrule
\end{tabular}
\label{tab:genaibench_results}
\end{table}

\section{Conclusion}
In this paper, we introduced GenAI-Arena, an open platform designed to rank generative models across text-to-image, image editing, and text-to-video tasks based on user preference. unlike other platforms, GenAI-Arena is driven by community voting to ensure transparency and sustainable operation. We employed the side-by-side human voting method to evaluate the models and collected over 9000 votes starting from February 11th, 2024. We compiled an Elo leaderboard with the votings and found that PlayGround V2.5, MagicBrush, and StableVideoDiffusion are the current state-of-the-art models in the three tasks (until Oct 24th, 2024). Analysis based on the collected votes shows that while the Elo rating is generally functional, but can biased by the imbalance of the "easy games" and "hard games". 
Our expert review of 350 sampled human votes confirmed that 93.07\% of the votes can be viewed as either clearly reasonable or vaguely reasonable, demonstrating the high quality of our collected votes
What's more, we also released the human preference voting as GenAI-Bench. We prompt the existing MLLMs to evaluate the generated images and videos on GenAI-Bench and compute the accuracy with human voting. The experiment showed that the open-source MLLMs achieve very low performance, even the best model GPT-4o can only achieve $49.19\%$ accuracy. This is mostly because their lack of instruction-following and reasoning ability in complex vision scenarios. In the future, we will continue collecting human votes to update the leaderboard, helping the community to keep track of the research progress. We also plan to develop a more robust MLLM to better approximate human ratings in GenAI-Bench. 

\clearpage

\bibliographystyle{abbrvnat}
\bibliography{nips_conference}

\newpage
\appendix

\section{Appendix}
\label{appendix}

\subsection{Broader Society Impacts}
\label{sec:impact}
The establishment of \arena and the release of GenAI-Bench have broader societal implications. By democratizing the evaluation of generative models, \arena encourages transparency and community engagement in AI development. This can lead to more trust in AI technologies as the public can gain insights into how models perform according to peer evaluations. Moreover, involving the community in such evaluations can accelerate the identification of potentially harmful biases or unethical uses of AI technologies. However, there are potential risks associated with the widespread use of generative AI technologies that \arena evaluates. For instance, advancements in text-to-image and text-to-video generation can be misused for creating misleading or harmful content, such as those filtered by NSFW Filter.

\subsection{Limitation}
\label{sec:limit}
While the release of \arena can enable a more reasonable evaluation of the generative models, there are several limitations in its development. First, the diversity and representativeness of the user base participating in \arena may not fully encapsulate the broader population's preferences, which will potentially bias the evaluation results. Despite efforts to attract voters with diverse backgrounds, there is an inherent challenge in ensuring a balanced representation across different cultures or professional backgrounds. In addition, the reliance on user feedback and votes introduces subjectivity into the evaluation process. While this is partially mitigated by the volume of data collected, individual biases and varying levels of expertise among users can skew the results. 

\subsection{Data Collection}
\label{sec:data_collect}
We stated in the \arena UI that the input and votes will be collected for research purposes only. By using this \arena tool, the users agree to the collection of their input and votes for research purposes. The users are acknowledged that their data will be anonymized and will not be used for commercial purposes.

\subsection{Extra Visualization on GenAI-Arena}
\begin{figure}[!ht]
    \centering
    \vspace{-1em}
    \begin{subfigure}[b]{0.32\textwidth}
        \centering
        \caption{Text-to-Image}
        \includegraphics[width=\textwidth]{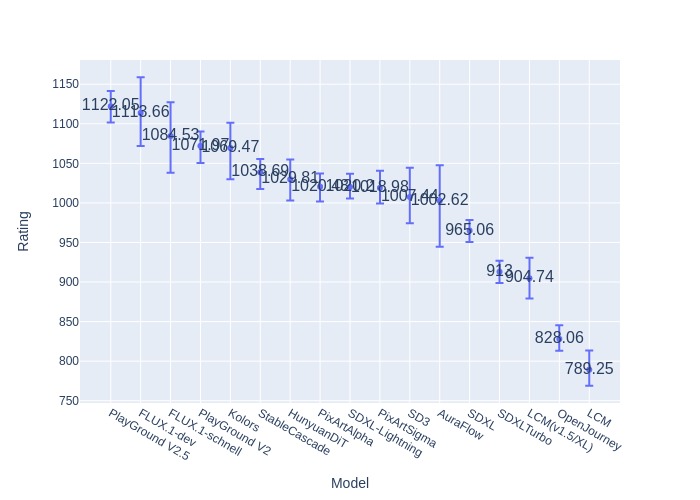}
        \label{fig:t2i_bootstrap}
    \end{subfigure}
    \hfill
    \begin{subfigure}[b]{0.32\textwidth}
        \centering
        \caption{Image Editing}
        \includegraphics[width=\textwidth]{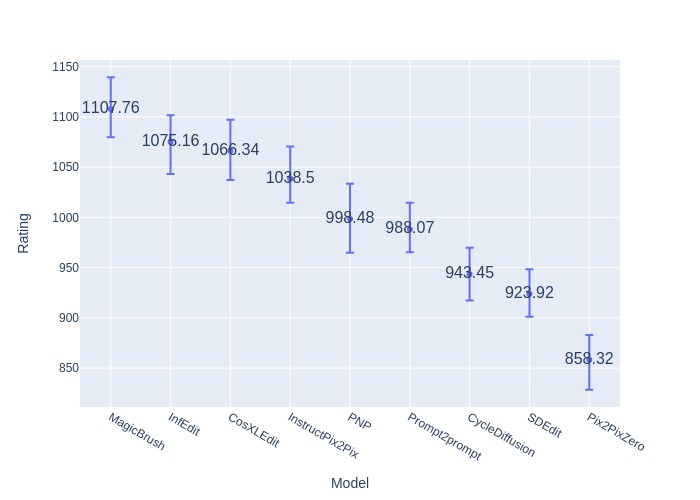}
        \label{fig:i2i_bootstrap}
    \end{subfigure}
    \hfill
    \begin{subfigure}[b]{0.32\textwidth}
        \centering
        \caption{Text-to-Video}
        \includegraphics[width=\textwidth]{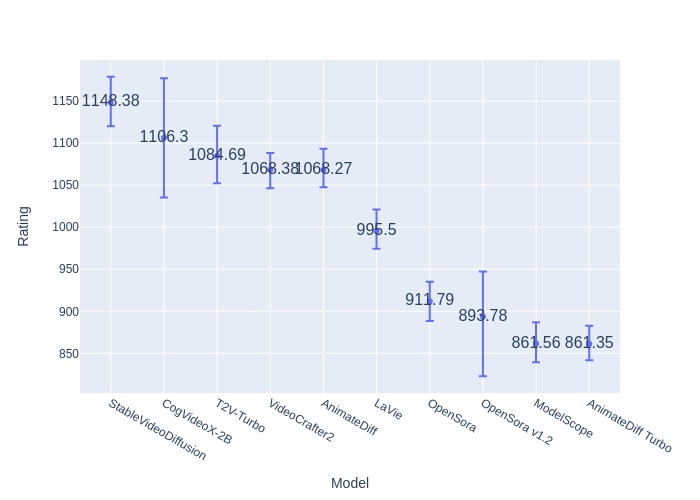}
        \label{fig:t2v_bootstrap}
    \end{subfigure}
    \vspace{-2em}
    \caption{Bootstrap of Elo Estimates (1000 Rounds of Random Sampling)}
    \label{fig:bootstrap}
\end{figure}

We included more analysis in Figure~\ref{fig:bootstrap} and~\ref{fig:avg_winrate} to show the reliability of GenAI-Arena. Specifically, Figure~\ref{fig:bootstrap} shows the error bar of the Elo rating to prove the reliability. For Figure~\ref{fig:avg_winrate}, it predicts the average win rate if the model is played against other models.

\subsection{VideoGenHub}
\label{sec:videogenhub}
VideoGenHub is an open-source library to standardize the inference and evaluation of all the conditional video generation models, similar to ImagenHub~\cite{ku2024imagenhub} in the image domain. In the library, all models are implemented with the literature standard, and the seeds are set as 42 for a fair comparison, which is the same standard as ImagenHub~\cite{ku2024imagenhub} implementation.

\clearpage
\subsection{Prompt Templates for GenAI-Bench}
\label{subsec:genai_bench_prompt_templates}
We provide the prompt templates used to prompt MLLM to output their preferences for the genai bench data in the followings. MLLMs are required to output 4 labels including \textsc{[[A>B]]}, \textsc{[[B>A]]}, \textsc{[[A=B=Good]]}, and \textsc{[[A=B=Bad]]}. Videos are extracted into image frames and fed into them as an image sequence, or directly fed into the model if the model have a specific video processing unit. We then compare their output labels with the real-world users preferences collected from out GenAI-Arena to judge a MLLM's ability in judging the quality of AI generative contents.

For text-to-image generation task, the prompt is as follows:
\begin{tcolorbox}
Please act as an impartial judge and a professional digital artist to evaluate the quality of the responses provided by two AI image generation models to the user inputs displayed below. You will be given model A's generated image and model B's generated image. Your job is to evaluate which assistant's generated image is better.\\
\\
Text prompt: <prompt>\\
Model A Generated Image: <left\_image>\\
Model B Generated Image: <right\_image>\\
\\
When evaluating the quality of the generated images, you must identify the any inappropriateness in the edited images by considering the following criteria:\\
1. Whether the text prompt has been followed successfully in the generated image.\\
2. Whether the generated image looks natural, such as the sense of distance, shadow, and lighting.\\
3. Whether the generated image contains any artifacts, such as distortion, watermark, scratches, blurred faces, unusual body parts, or subjects not harmonized.\\
4. Whether the generated image is visually appealing and esthetically pleasing.\\
\\
After providing your explanation, you must output only one of the following choices as your final verdict with a label:\\
\\
1. Model A is better: [[A>B]]\\
2. Model B is better: [[B>A]]\\
3. Tie, relatively the same acceptable quality: [[A=B=Good]]\\
4. Both are bad: [[A=B=Bad]]\\
\label{box:genai_bench_image_generation_prompt}
\end{tcolorbox}

\clearpage
For image-edition task, the prompt is as follows:
\begin{tcolorbox}
Please act as an impartial judge and a professional digital artist to evaluate the quality of the responses provided by two AI image edition models to the user inputs displayed below. You will be given model A's edited image and model B's edited image. Your job is to evaluate which assistant's edited image is better.\\
\\
Source Image prompt: <source\_prompt>\\
Target Image prompt after editing: <target\_prompt>\\
Editing instruction: <instruct\_prompt>\\
Source Image: <source\_image>\\
\\
Model A Edited Image: <left\_output\_image>\\
Model B Edited Image: <right\_output\_image>\\
\\
When evaluating the quality of the edited images, you must identify the any inappropriateness in the edited images by considering the following criteria:\\
1. Whether the editing instruction has been followed successfully in the edited image.\\
2. Whether the edited image is overedited, such as the scene in the edited image is completely different from the original.\\
3. Whether the edited image looks natural, such as the sense of distance, shadow, and lighting.\\
4. Whether the edited image contains any artifacts, such as distortion, watermark, scratches, blurred faces, unusual body parts, or subjects not harmonized.\\
5. Whether the edited image is visually appealing and esthetically pleasing.\\
\\
After providing your explanation, you must output only one of the following choices as your final verdict with a label:\\
\\
1. Model A is better: [[A>B]]\\
2. Model B is better: [[B>A]]\\
3. Tie, relatively the same acceptable quality: [[A=B=Good]]\\
4. Both are bad: [[A=B=Bad]]\\
\label{box:genai_bench_image_edition_prompt}
\end{tcolorbox}

\clearpage
For video-generation tasks, the prompt is as follows:
\begin{tcolorbox}
Please act as an impartial judge and a professional digital artist to evaluate the quality of the responses provided by two AI video generation models to the user inputs displayed below. You will be given model A's generated video and model B's generated video. Your job is to evaluate which assistant's generated video is better.\\
\\
Text prompt: <prompt>\\
Model A Generated Video: <left\_video>\\
Model B Generated Video: <right\_video>\\
\\
When evaluating the quality of the generated videos, you must identify the any inappropriateness in the edited videos by considering the following criteria:\\
1. Whether the text prompt has been followed successfully in the generated video.\\
2. Whether the generated video looks natural, such as the sense of distance, shadow, and lighting.\\
3. Whether the generated video is good visual quality, such as clearness, resolution, brightness, and color.\\
4. Whether the generated video is consistent and coherent in terms of the scene, objects, and characters.\\
5. Whether the generated video is dynamic and not static like a single image.\\
6. Whether the generated video is visually appealing and esthetically pleasing.\\
\\
After providing your explanation, you must output only one of the following choices as your final verdict with a label:\\
\\
1. Model A is better: [[A>B]]\\
2. Model B is better: [[B>A]]\\
3. Tie, relatively the same acceptable quality: [[A=B=Good]]\\
4. Both are bad: [[A=B=Bad]]\\
\label{box:genai_bench_video_generation_prompt}
\end{tcolorbox}



\end{document}